\def\year{2019}\relax
\documentclass[letterpaper]{article} 
\usepackage{aaai19}  
\usepackage{times}  
\usepackage{helvet}  
\usepackage{courier}  
\usepackage{url} 
\usepackage{graphicx}  
\usepackage{amssymb}
\usepackage{algorithm}
\usepackage[noend]{algpseudocode}
\usepackage{algorithmicx}
\usepackage{booktabs} 
\usepackage{graphicx}
\usepackage{amsmath,amsthm,mathrsfs,bm}
\usepackage{amsfonts}  
\usepackage{amssymb}  
\usepackage{etoolbox}
\usepackage{array}
\usepackage[inline]{enumitem}
\usepackage{subfigure}
\usepackage{tikz}  
\usepackage{multirow}

\makeatletter
\newcommand{\algmargin}{\the\ALG@thistlm}
\makeatother

\algnewcommand{\parState}[1]{\State%
	\parbox[t]{\dimexpr\linewidth-\algmargin}{\strut #1\strut}}

\newtheorem{thm}{Theorem}
\newtheorem{dfn}{Definition}

\newtheorem{lem}{Lemma}

\newtheorem{ex}{Example}
\usepackage{thm-restate}

\AtEndEnvironment{ex}{}

\title{Practical Algorithms for Multi-Stage Voting Rules\\
 with Parallel Universes Tiebreaking}
\author{Jun Wang\and Sujoy Sikdar\and Tyler Shepherd\\
	 {\bf \Large Zhibing Zhao\and Chunheng Jiang \and Lirong Xia}\\
	Rensselaer Polytechnic Institute\\
	110 8th Street, Troy, NY, USA\\
	\{wangj38, sikdas, shepht2, zhaoz6, jiangc4\}@rpi.edu, xial@cs.rpi.edu\\
}

\begin{document}
%
\maketitle
\begin{abstract}
STV and ranked pairs (RP) are two well-studied voting rules for group decision-making. They proceed in multiple rounds, and are affected by how ties are broken in each round. However, the literature is surprisingly vague about how ties should be broken. We propose the first algorithms for computing the set of alternatives that are winners under {\em some} tiebreaking mechanism under STV and RP, which is also known as parallel-universes tiebreaking (PUT). Unfortunately, {\em PUT-winners} are NP-complete to compute under STV and RP, and standard search algorithms from AI do not apply. We propose multiple DFS-based algorithms along with pruning strategies, heuristics, sampling and machine learning to prioritize search direction to significantly improve the performance. We also propose novel ILP formulations for PUT-winners under STV and RP, respectively. Experiments on synthetic and real-world data show that our algorithms are overall faster than ILP.
\end{abstract}

\section{Introduction}
The {\em Single Transferable Vote (STV)} rule\footnote{STV is also known as {\em instant runoff voting, alternative vote}, or {\em ranked choice voting}.} is among the most popular voting rules used in real-world elections. According to Wikipedia, STV is being used to elect senators in Australia, city councils in San Francisco (CA, USA) and Cambridge (MA, USA), and more \cite{Wikipedia:STV}. In each round of STV, the lowest preferred alternative is eliminated, in the end leaving only one alternative, the winner, remaining.

This raises the question: {\em when two or more alternatives are tied for last place, how should we break ties to eliminate an alternative?} The literature provides no clear answer. For example, see~\cite{stvvariants} for a list of different STV tiebreaking variants. While the STV winner is unique and easy to compute for a fixed tiebreaking mechanism, it is NP-complete to compute all winners under {\em all} tiebreaking mechanisms. This way of defining winners is called parallel-universes tiebreaking (PUT) \cite{Conitzer09:Preference}, and we will therefore call them \textit{PUT-winners} in this paper.

Ties do actually occur in real-world votes under STV. On Preflib~\cite{Mattei13:Preflib} data, 9.2\% of profiles have more than one PUT-winner under STV. There are two main motivations for computing all PUT-winners. First, it is vital in a democracy that the outcome not be decided by an arbitrary or random tiebreaking rule, which will violate the {\em neutrality} of the system ~\cite{Brill12:Price}. Second, even for the case of a unique PUT-winner, it is important to prove that the winner is unique despite ambiguity in tiebreaking. In an election, we would prefer the results to be transparent about who all the winners could have been. 

A similar problem occurs in the Ranked Pairs (RP) rule, which satisfies many desirable axiomatic properties in social choice~\cite{Schulze11:New}. The RP procedure considers every pair of alternatives and builds a ranking by selecting the pairs with largest victory margins. This continues until every pair is evaluated, the winner being the candidate which is ranked above all others by this procedure \cite{Tideman87:Independence}. Like in STV, ties can occur, and the order in which pairs are evaluated can result in different winners.  Unfortunately, like STV, it is NP-complete to compute all PUT-winners under RP~\cite{Brill12:Price}.

More generally, the tiebreaking problem exists for a larger class of voting rules called {\em multi-stage voting rules}. These rules eliminate alternatives in multiple rounds, and the difference is in the elimination methods. For example, in each round, Baldwin's rule eliminates the alternative with the smallest Borda score, and Coombs eliminates the alternative with highest veto score. Like STV and RP, computing all PUT winners is NP-complete for these multi-stage rules~\cite{Mattei2014:How-hard}.

To the best of our knowledge, no algorithm beyond brute-force search is known for computing PUT-winners under STV, RP, and other multi-stage voting rules.  Given its importance as discussed above, the question we address in this paper is:

 {\em How can we design efficient, practical algorithms for computing PUT-winners under multi-stage voting rules?}

~\\
\noindent{\bf Our Contributions.}
Our main contributions are the first practical algorithms to compute the PUT-winners for multi-stage voting rules: a depth-first-search (DFS) framework and integer linear programming (ILP) formulations. 

In our DFS framework, the nodes in the search tree represent intermediate rounds in the multi-stage rule, each leaf node is labeled with a single winner, and each root-to-leaf path represents a way to break ties. The goal is to output the union set of winners on the leaves. See Figure~\ref{fig:stvex} and Figure~\ref{fig:rpex} for examples. To improve the efficiency of the algorithms, we propose the following techniques: 

~\\
\noindent{\bf Pruning}, which maintains a set of {\em known winners} during the search procedure and can then prune a branch if expanding a state can never lead to any new PUT-winner.

~\\
\noindent{\bf Machine-Learning-Based Prioritization}, which aims at building a large known winner set as soon as possible by prioritizing nodes that minimize the number of steps to discover a new PUT-winner.

~\\
\noindent{\bf Sampling}, which build a large set of known winners before the search to make it easier to trigger the pruning conditions.

Our main conceptual contribution is a new measure called {\em early discovery}, wherein we time how long it takes to compute a given proportion of all PUT-winners on average. This is particularly important for low stakes and anytime applications, where we want to discover as many PUT-winners as possible  at any point during execution.

We will use STV and RP in this paper as illustrations for our framework. Experiments on synthetic and real-world data show the efficiency and effectiveness of our algorithms in solving the PUT problem for STV and RP, hereby denoted PUT-STV and PUT-RP respectively. The effects of additional techniques compared to the standard DFS framework are summarized in Table~\ref{tbl:contribution}, where the symbol $++$ denotes very useful, $+$ denotes mildly useful, and 0 denotes not useful.
\begin{table}[htp]
	\centering	
	\begin{tabular}[width=\linewidth]{|r|c|c|}\hline
		&PUT-STV &PUT-RP\\\hline
		Pruning & $+$  &$++$\\ \hline
		Machine Learning & $+$ & $+$\\ \hline
		Sampling & 0 & $++$ \\\hline		
	\end{tabular}
	\caption{\label{tbl:contribution}Summary of Technique Effectiveness.}
\end{table}

It turns out that the standard DFS algorithm is already efficient for STV, while various new techniques significantly improve running time and early discovery for RP.

In addition, we design an ILP for STV, and an ILP for RP based on the characterization by Zavist and Tideman (1989). For both PUT-STV and PUT-RP, in the large majority of cases our DFS-based algorithms are orders of magnitude faster than solving the ILP formulations, but there are a few cases where ILP for PUT-RP is significantly faster. This means that both types of algorithm have value and may work better for different datasets.

\noindent{\bf Related Work and Discussions.}
There is a large literature on the computational complexity of winner determination under commonly-studied voting rules. In particular, computing winners of the Kemeny rule has attracted much attention from researchers in AI and theory \cite{Conitzer06:Kemeny,Kenyon07:How}. However, STV and ranked pairs have both been overlooked in the literature, despite their popularity. We are not aware of previous work on practical algorithms for PUT-STV or PUT-RP. 
A recent work on computing winners of commonly-studied voting rules proved that computing STV is P-complete, but only with a fixed-order tiebreaking mechanism \cite{Csar2017:Winner}. Our paper focuses on finding all PUT-winners under all tiebreaking mechanisms. See \cite{Freeman2015:General} for more discussions on tiebreaking mechanisms in social choice. 

Standard procedures to AI search problems unfortunately do not apply here. In a standard AI search problem, the goal is to find a path from the root to the goal state in the search space. However, for PUT problems, due to the unknown number of PUT-winners, we do not have a clear predetermined goal state. Other voting rules, such as Coombs and Baldwin, have similarly been found to be NP-complete to compute PUT winners~\cite{Mattei2014:How-hard}. The techniques we apply in this paper for STV and RP can be extended to these other rules, with slight modification based on details of the rules. 

\section{Preliminaries}\label{sec:prelim}
Let $\mathcal A=\{a_1,\cdots, a_m\}$ denote a set of $m$ {\em alternatives} and let $\mathcal L(\mathcal A)$ denote the set of all possible linear orders over $\mathcal A$. A {\em profile} of $n$ voters is a collection $P =(V_i)_{i\le n}$ of votes where for each $i\leq n$, $V_i\in \mathcal L(\mathcal A)$. A voting rule takes as input a profile and outputs a non-empty set of winning alternatives. 

\noindent{\bf Single Transferable Vote (STV)}
proceeds in $m-1$ rounds over alternatives $\mathcal A$ as follows. In each round, \begin{enumerate*}[label=(\arabic*)] \item an alternative with the lowest plurality score is eliminated, and \item the votes over the remaining alternatives are determined.\end{enumerate*} The last-remaining alternative is  the winner.

\begin{figure}[H]
	\includegraphics[width=1\linewidth]{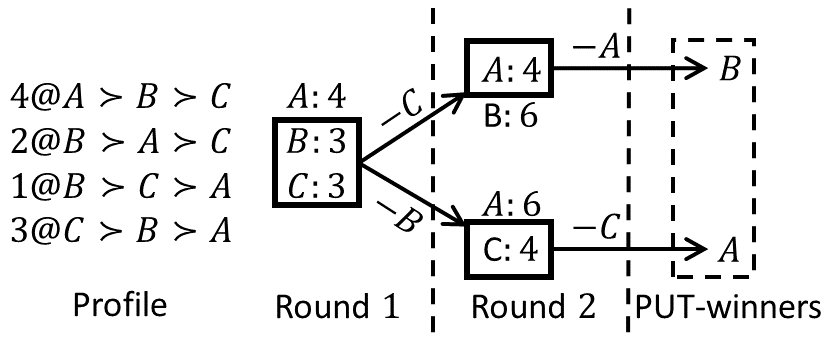}
	\caption{\label{fig:stvex} An example of the STV procedure.}
\end{figure}

\begin{ex}
	Figure~\ref{fig:stvex} shows an example of how the STV procedure can lead to different winners depending on the tiebreaking rule. In round 1, alternatives $B$ and $C$ are tied for last place. For any tiebreaking rule in which $C$ is eliminated, this leads to $B$ being the winner. Alternatively, if $B$ were to be eliminated, then $A$ is the winner.
\end{ex}

\noindent{\bf Ranked Pairs (RP)}. For a given profile $P=(V_i)_{i\le n}$, we define the 

{\em weighted majority graph} (WMG) of $P$, denoted by wmg$(P)$, to be the weighted digraph $(\mathcal A,E)$ where the nodes are the alternatives, and for every pair of alternatives $a,b \in \mathcal A$, there is an edge $(a,b)$ in $E$ with weight $w_{(a,b)} = |\{V_i: a \succ_{V_i} b\}| - |\{V_i: b \succ_{V_i} a\}|$. We define the \textit{nonnegative WMG} as wmg$_{\geq 0}(P) = (\mathcal A, \{(a,b) \in E: w_{(a,b)} \ge 0\})$. We partition the edges of wmg$_{\geq 0}(P)$ into tiers $T_{1},...,T_{K}$ of edges, each with distinct edge weight values, and indexed according to decreasing value. Every edge in a tier $T_i$ has the same weight, and for any pair $i,j\le n$, if $i < j$, then $\forall e_1 \in T_i, e_2 \in T_j, w_{e_1} > w_{e_2}$.

\begin{figure}[h]
	\includegraphics[width=0.95\linewidth]{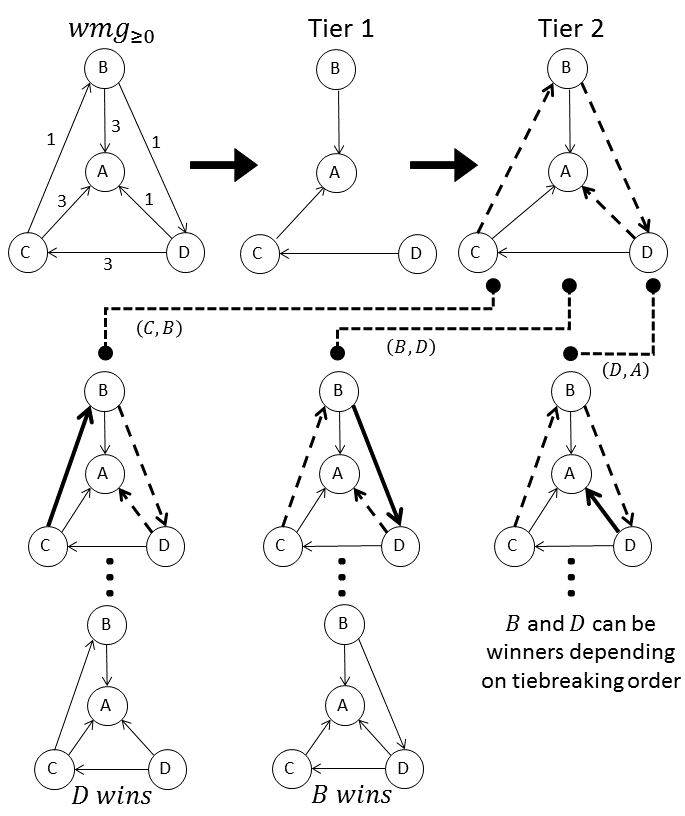}
	\caption{\label{fig:rpex} An example of the RP procedure.}
\end{figure}

Ranked pairs proceeds in $K$ rounds: Start with an empty graph $G$ whose vertices are $\mathcal A$. 
In each round $i \le K$, consider adding edges $e \in T_i$ to $G$ one by one according to a tiebreaking mechanism, as long as it does not introduce a cycle. 
Finally, output the top-ranked alternative in $G$ as the winner.

\begin{ex}
	Figure~\ref{fig:rpex} shows the ranked pairs procedure applied to the WMG resulting from a profile over $m = 4$ alternatives (a profile with such a WMG always exists) \cite{McGarvey53:Theorem}. We focus on the addition of edges in tier $2$, where $\{(C,B) ,(B,D),(D,A)\}$ are to be added. Note that $D$ is the winner if $(C,B)$ is added first , while $B$ is the winner if $(B,D)$ is added first. 
\end{ex}

\section{General Framework}\label{sec:greneral}
We provide a general framework using a depth first search approach to solve multi-stage voting rules in Algorithm~\ref{algo:dfs}. Our algorithms for solving PUT-STV and PUT-RP are examples of applying this common framework.

\begin{algorithm}[ht]
	\begin{algorithmic}[1]
	\State{\bf Input:} A profile $P$.
	\State{\bf Output:} All PUT-winners $W$.
	\parState{Initialize a stack $F$ with the initial state; $W=\varnothing$.}
	\parState{\textit{Sample} PUT-winners randomly. \label{step:dfs_sample}}
	\While{$F$ is not empty}\label{step:while}
		\State Pop a state $S$ from $F$ to explore.
		\If{$S$ has a single remaining alternative}
			\State add it to $W$.
		\EndIf
		\If{$S$ already visited or state can be \textit{pruned}} \label{step:dfs_prune}
			\State skip state.
		\EndIf
		\parState{Expand current state to generate children $C$, add $C$ to $F$ in order of \textit{priority}. \label{step:dfs_expand}}
	\EndWhile
	\State \Return $W$.
	\end{algorithmic}
	\caption{General DFS-based framework. \label{algo:dfs}}
\end{algorithm}
We evaluate our algorithms on two criteria: 
\begin{enumerate}[label=(\arabic*)]
	\item {\bf Total Running Time} for completing the search.
	\item {\bf Early Discovery.} For any PUT-winner algorithm and any number $0\le \alpha\le 1$, the $\alpha$-discovery value is the average running time for the algorithm to compute $\alpha$ fraction of PUT-winners.
\end{enumerate}

The early discovery property is motivated by the need for {\em any-time} algorithms which, if terminated at any time, can output the known PUT-winners as an approximation to all PUT-winners. Such time constraint is realistic in low-stakes, everyday voting systems such as Pnyx~\cite{Brandt2015:Pnyx}, and it is desirable that an algorithm outputs as many PUT-winners as early as possible. This is why we focus on DFS-based algorithms, as opposed to, for example, BFS, since the former reaches leaf nodes faster. We note that $100\%$-discovery value can be much smaller than the total running time of the algorithm, because the algorithm may continue exploring remaining nodes after $100\%$ discovery to verify that no new PUT-winner exists.

We propose several techniques to improve the running time and guide the search into promising branches containing more alternatives that could be PUT-winners.

~\\
\noindent{\bf Pruning.} The main idea is straightforward: if all candidate PUT-winners of the current state are already known to be PUT-winners, then there is no need to continue exploring this branch since no new PUT-winners can be found.

\noindent{\bf Heuristic Functions.}  
To achieve early discovery through Algorithm~\ref{algo:dfs}, we prioritize nodes whose state contains more candidate PUT-winners that have not been discovered. As a heuristic guided approach, we devise {\em local priority} functions which take as input the set of candidate PUT-winners (denoted by $A$) and the set of known PUT-winners (previously discovered by the search, denoted by $W$) and output a priority order on the branches to explore in Line~\ref{step:dfs_expand} of Algorithm~\ref{algo:dfs}. It is called local (as opposed to global) priority because overall the algorithm is still DFS, and the local priority is used to decide which child of the current node will be expanded first.

\noindent$\bullet$~$\text{LP} = |A - W|$: Local priority; orders the exploration of children by the value of $|A - W|$, the number of potentially unknown PUT-winners.

\noindent{\bf Machine Learning.}  We propose to use machine learning as a novel heuristic to guide the search by estimating the probability of a branch to have new PUT-winners, and prioritizing the exploration of branches with a higher estimated probability of having new PUT-winners.

\noindent$\bullet$~$\text{LPML} = \sum_{a \in A-W} \pi(a)$: Local priority with machine learning model $\pi$.

Here $\pi(a)$ is the machine learning model probability of $a$ to be a PUT-winner. The setup details can be found in Section~\ref{sec:results}. It is important to note we do \textit{not} use the machine learning model to directly predict PUT-winners. Instead, we use it to guide DFS search to discover a new PUT-winner as soon as possible.

\noindent{\bf Sampling.} Sampling in line~\ref{step:dfs_sample} of Algorithm~\ref{algo:dfs} can be seen as a preprocessing step: before running the search, we repeatedly randomly sample a fixed tie-breaking order $\pi$ and run the voting procedure using $\pi$. If we can add PUT-winners earlier into the known winners set, the algorithm will earlier reach the pruning conditions during the search, eliminating branches.

{\noindent\bf PUT-STV as an Example.} The framework in Algorithm~\ref{algo:dfs} can be applied to PUT-STV with the following specifications. We set the initial state in $F$ to be the set of all alternatives $\mathcal A$. We modify Step~\ref{step:dfs_expand} of Algorithm~\ref{algo:dfs} as: for every remaining lowest plurality score alternative $c \in S$, in order of \textit{priority} add ($S \setminus c$) to $F$. For pruning specific to PUT-STV, we can skip the state $S$ whenever $S \subseteq W$. We implement the heuristic functions LP and LPML to prioritize the order of exploring children, where the set of candidate PUT-winners $A$ is simply the remaining candidates of state $S$. And we finally add sampling to our algorithms. The results are in Section~\ref{sec:results}.

\section{Algorithms for PUT-RP}\label{sec:rppractical}
In this section we show how to adopt Algorithm~\ref{algo:dfs} to compute PUT-RP.  In the search tree, each node has a state $(G,E)$, where $E$ is the set of edges that have not been considered yet and $G$ is a graph whose edges are pairs that have been ``locked in'' by the RP procedure according to some tiebreaking mechanism. The root node is $(G = (\mathcal A, \varnothing),E_0)$, where $E_0$ is the set of edges in wmg$_{\ge0}(P)$.

At a high level, we have two ways to apply Algorithm~\ref{algo:dfs}. The first one is called {\em naive DFS} because it is a straightforward application of Algorithm~\ref{algo:dfs}, generating children at state $(G,E)$ by adding each edge in the highest weight edge tier of $E$ to $G$ that does not cause a cycle. We also include our pruning conditions as detailed in this section.

The second method is, in short, a layered algorithm in which we process edges tier by tier. It is described as the $PUT$-$RP()$ procedure in Algorithm~\ref{algo:rpdfs}.  
Exploring a node $(G,E)$ at depth $t$ involves finding all maximal ways of adding edges from $T_t$ to $G$ without causing a cycle, which is done by the $MaxChildren()$ procedure shown in Algorithm~\ref{algo:maxchild}. $MaxChildren()$ takes a graph $G$ and a set of edges $T$ as input, and follows a DFS-like addition of edges one at a time. Within the algorithm, each node $(H,S)$ at depth $d$ corresponds to the addition of $d$ edges from $T$ to $H$ according to some tiebreaking mechanism. $S \subseteq T$ is the set of edges not considered yet.

\begin{dfn}
	Given a directed acyclic graph $G = (\mathcal A,E)$, and a set of edges $T$, a graph $C=(\mathcal A, E\cup T')$ where $T'\subseteq T$ is a {\em maximal child} of $(G,T)$ if and only if $\forall e\in T\backslash T'$, adding $e$ to the edges of $C$ creates a cyclic graph. 
\end{dfn}

\begin{algorithm}[ht]
	\begin{algorithmic}[1]
	\State {\bf Input:} A profile $P$. 
	\State {\bf Output:} All PUT-RP winners $W$.
	\State Compute $(\mathcal A, E_0) = \text{wmg}_{\ge 0}(P)$
	\parState{Initialize a stack $F$ with $((\mathcal A,\varnothing),E_0)$ for DFS; $W = \varnothing$.}
	\While {$F$ is not empty}
		\State 	Pop a state $(G, E)$ from $F$ to explore.
		\If{$E$ is empty or this state can be pruned}
			\parState{Add all topologically top vertices of $G$ to $W$ and skip state.}
		\EndIf
		\State $T \gets$ highest tier edges of $E$.
		\For{$C$ in $MaxChildren(G,T)$}
			\State Add $(C, E \setminus T)$ to $F$.
		\EndFor
	\EndWhile
	\State \Return $W$.
	\end{algorithmic}
	\caption{PUT-RP$(P)$\label{algo:rpdfs}}
	
\end{algorithm}

\begin{algorithm}[ht]
	\begin{algorithmic}[1]
	\State {\bf Input:} A graph $G=(\mathcal A,E)$, and a set of edges $T$.
	\State {\bf Output:} Set $C$ of all maximal children of $G, T$.
	\State Initialize a stack $I$ with $(G,T)$ for DFS; $C = \varnothing$.
	\While{$I$ is not empty}
		\State Pop $((\mathcal A,E'),S)$ from $I$.
		\If{$E'$ already visited or state can be \textit{pruned}}
			\State skip state.
		\EndIf
		\parState{The successor states are $Q_e = (G_e,S\setminus{e})$ for each edge $e$ in $S$, where graph $G_e = (\mathcal A, E' + e)$.}
		\State Discard states where $G_e$ is cyclic.
		\If{in all successor states $G_e$ is cyclic}
			\parState{We have found a max child; add $(\mathcal A, E')$ to $C$.}
		\Else
			\parState{Add states $Q_e$ to $I$ in order of \textit{local priority}.}
		\EndIf
	\EndWhile
	\State \Return $C$.
	\caption{MaxChildren$(G,T)$}
	\label{algo:maxchild}
	\end{algorithmic}
\end{algorithm}

Therefore, the second algorithm will be called maximal children based (MC) algorithms. We have the following techniques for PUT-RP.

\noindent{\bf Pruning.}
For a graph $G$ and a tier of edges $T$, we implement the following conditions to check if we can terminate exploration of a branch of DFS early: \begin{enumerate*}[label=(\roman*)] \item If every alternative that is not a known winner has one or more incoming edges or \item If all but one vertices in $G$ have indegree $> 0$, the remaining alternative is a PUT-winner. \end{enumerate*} 
For example, in Figure~\ref{fig:rpex}, we can prune the right-most branch after having explored the two branches to its left.

\noindent{\bf SCC Decomposition.} We further improve Algorithm~\ref{algo:maxchild} by computing strongly connected components (SCCs). For a digraph, an SCC is a maximal subgraph of the digraph where for each ordered pair $u$, $v$ of vertices, there is a path from $u$ to $v$ . Every edge in an SCC is part of some cycle. The edges not in an SCC, therefore not part of any cycle, are called the bridge edges \cite[p.~98-99]{KleinbergEva:Algorithm}. 

Given a graph $G$ and a set of edges $T$, finding the maximal children will be simpler if we can split it into multiple SCCs. We find the maximal children of each SCC, then combine them in the Cartesian product with the maximal children of every other SCC. Finally, we add the bridge edges. 

Figure 3 shows an example of SCC Decomposition in which edges in $G$ are solid and edges in $T$ are dashed. Note this is only an example, and does not show all maximal children. In the unfortunate case when there is only one SCC we cannot apply SCC decomposition. The proof of Theorem~\ref{thm:scc} is provided in Appendices.

\begin{restatable}{thm}{mainclaim}\label{thm:scc}
	For any directed graph $H$, $C$ is a maximal child of $H$\footnote{Here, we extend the definition of maximal child of directed graph $H = (\mathcal A,E)$ as the  maximal child of  the tuple $(G,E)$ where $G=(\mathcal A,\varnothing)$.} if and only if $C$ contains exactly \begin{enumerate*}[label=(\roman*)]
		\item all bridge edges of $H$ and \item the union of the maximal children of all SCCs in $H$.
	\end{enumerate*}
\end{restatable}

\begin{figure}
	\centering
	\includegraphics[scale = 0.7]{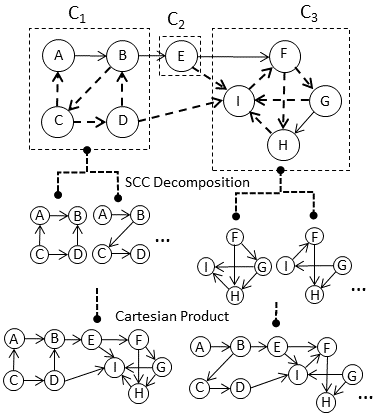}
	\caption{\label{signed graph_EP} Example of SCC Decomposition.} 
\end{figure}

\section{Experiment Results}\label{sec:results}

\subsection{Datasets}\label{sec:datasets}
We use both synthetic datasets and  real-world preference profiles from Preflib to test our algorithms' performance. The synthetic datasets were generated based on impartial culture with $n$ independent and identically distributed rankings uniformly at random over $m$ alternatives for each profile.  From the randomly generated profiles, we only test on \textit{hard} cases where the algorithm encounters a tie that cannot be solved through simple pruning. All the following experiments are completed on a PC with Intel i5-7400 CPU and 8GB of RAM running Python 3.5.

\begin{figure}[htp]
	\centering
	\includegraphics[width=0.8\linewidth]{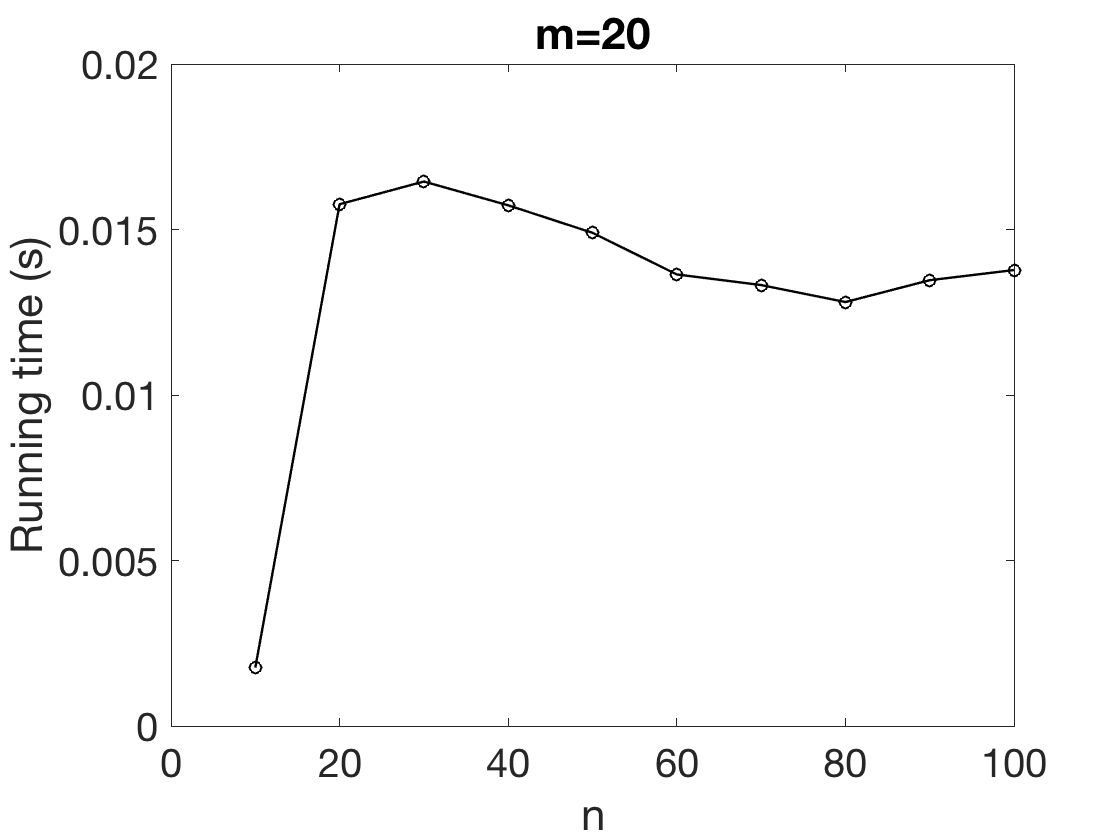}
	\caption{\label{P1P0} Running time of DFS for PUT-STV for different number of voters $n$, for profiles with $m=20$ candidates.}	
\end{figure}

~\\
\noindent{\bf Synthetic Data.} For PUT-STV, we generate 10,000 $m=n=30$ synthetic hard profiles. For PUT-RP, we generate 14,875 $m=n=10$ synthetic hard profiles. We let $m=n$ in our synthetic data because these are the hardest cases, which can be verified in Figure~\ref{P1P0}.\\

\noindent{\bf Preflib Data.} We use all available datasets on Preflib suitable for our experiments on both rules. Specifically, we use 315 profiles from Strict Order-Complete Lists (SOC), and 275 profiles from Strict Order-Incomplete Lists (SOI). They represent several real world settings including political elections, movies and sports competitions. For political elections, the number of candidates is often not more than 30. For example, 76.1\% of 136 SOI election profiles on Preflib has no more than 10 candidates, and 98.5\% have no more than 30 candidates.

\begin{table*}
	\centering
	\begin{tabular}[width=\linewidth]{|r|c|c|c|c|} \hline
		\multirow{ 2}{*}{}&\multicolumn{2}{c|}{without sampling}& \multicolumn{2}{c|}{with sampling}\\ \cline{2-5}
		&	DFS	&  LPML&	LPML+$600$ samples&LPML+$30$ samples\\ \hline
		Avg. running time (s)&	0.4474&		0.5017&	0.6893&	0.5252\\
		Avg. 100\%-discovery time (s) &	0.1820&	0.1686&	0.3563&	0.1826\\
		\hline
	\end{tabular}
	\caption{\label{table:STV} Experiment results of different algorithms for PUT-STV.}
\end{table*}

\subsection{Results for PUT-STV}\label{sec:stvresults}
\begin{figure}[]
	\centering
	\includegraphics[width=1\linewidth]{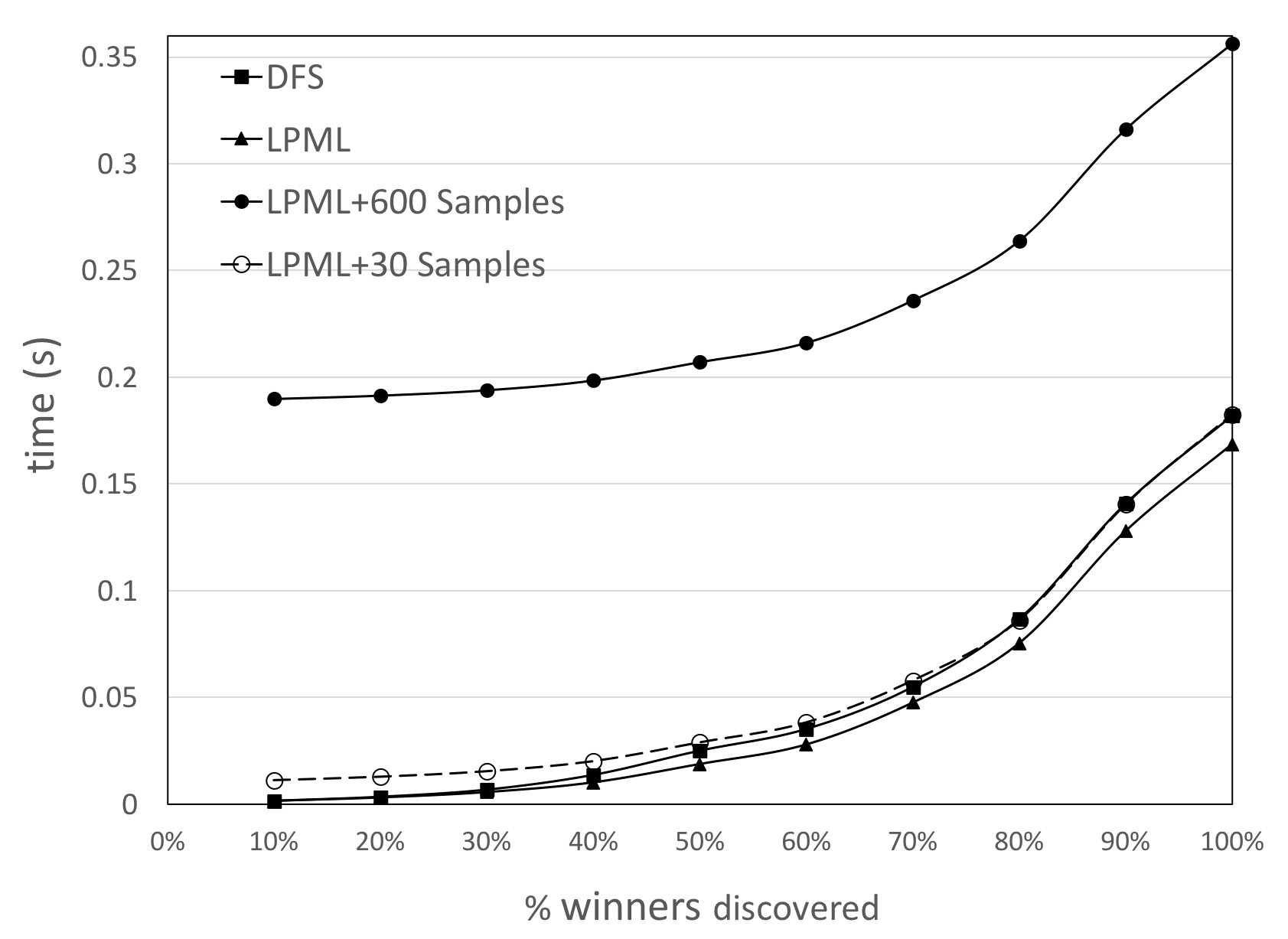}
	\caption{ PUT-STV early discovery. $m=30$.\label{fig:STV_discovery}}
\end{figure}

We test four variants of Algorithm~\ref{algo:dfs} for PUT-STV: \begin{enumerate*}[label=(\roman*)]\item {\bf DFS:} Algorithm~\ref{algo:dfs} with pruning but without local priority , \item  {\bf LPML:} Algorithm~\ref{algo:dfs} with pruning and local priority based on machine learning,  \item  {\bf LPML+$\bm{20m}$ samples:} its variants with $20m=600$ samples, and \item {\bf LPML+$\bm{m}$ samples:}  $m=30$ samples.\end{enumerate*} 
Results are summarized in Table~\ref{table:STV} and Figure~\ref{fig:STV_discovery}. The main conclusions are that DFS performs the best in total running time, local priority based on machine learning (LPML) is useful for early discovery, and sampling is not useful. This makes sense because expanding nodes for STV is computationally easy, so the operational cost of computing and maintaining a priority queue in LPML offsets the benefit of early discovery. All comparisons are statistically significant with p-value=0.02 or less computed using one sided paired sample Student’s t-test.

~\\
\noindent{\bf Local Priority with Machine Learning Improves Early Discovery.}\label{sec:stvheuristic} As shown in Figure~\ref{fig:STV_discovery}, for $m=n=30$,  
LPML has $25.01\%$ reduction in 50\%-discovery compared to DFS. Results are similar for other datasets with different $m$. The early discovery figure is computed by averaging the time to compute a given percentage $p$ of PUT-winners. For example, for a profile with 2 PUT-winners which are discovered at time $t_1$ and $t_2$, we set the 10\%-50\% discovery time as $t_1$ and the 60\%-100\%-discovery time as $t_2$.

The $\pi$ function used in the local priority function was trained by a neural network model using three hidden layers with size of $4096\times 1024\times 1024$ neurons and logistic function as activation, where the output has $m$ components, each of which indicates whether the corresponding alternative is a PUT-winner. The input features are the positional matrix, WMG, and plurality, Borda, k-approval, Copeland and maximin scores of the alternatives. We trained the models on 50,000 $m=n=30$ hard profiles using tenfold cross validation, with the objective of minimizing the $L1$-distance between the prediction vector and the target true winner vector. Our mean squared error was 0.0833.

\noindent{\bf Pruning Has Small Improvement.}\label{sec:stvcp} We see only a small improvement in the running time when evaluating pruning: on average, pruning brings only 0.33\% reduction in running time for $m=n=10$ profiles, 2.26\% for $m=n=20$ profiles, and 4.51\% for $m=n=30$ profiles.

\noindent{\bf Sampling Does Not Help.}\label{sec:stvsampl} We test different number of samples as shown in Figure~\ref{fig:STV_discovery} but none of them brings improvement. This is because each sample is essentially just a run of DFS up to leaf node, which is no different from our algorithm. Moreover, pruning has only small improvement since its condition is not often triggered, so knowing PUT-winners actually does no good to both early discovery and running time.

\noindent{\bf DFS is Practical on Real-World Data.} Our experimental results on Preflib data show that on 315 complete-order real world profiles, the maximum observed running time is only $0.06$ seconds and the average is 0.335 ms.

\begin{table*}
	\centering
	\begin{tabular}[width=\linewidth]{|r|c|c|c|c|c|c|} \hline
		
		\multirow{ 2}{*}{}&\multicolumn{3}{c|}{without sampling}& \multicolumn{3}{c|}{with sampling}\\ \cline{2-7}
		&	nDFS(LP)	&  MC(LPML)&	MC(LP)&nDFS(LP)	&MC(LPML)&MC(LP)\\ \hline
		Avg. running time (s)&	7.6571&		7.9858&	7.7081&7.5291&	3.0395&	2.8692\\
		Avg. 100\%-discovery time (s) &	0.2778&		6.7876&	6.4058&	0.0531&0.2273&	0.0823\\
		\hline
	\end{tabular}
	\caption{\label{table:RP} Experiment results for PUT-RP.}
\end{table*}

\subsection{PUT-RP}\label{sec:rpresults} 
\begin{figure}[]
	\centering
	\includegraphics[width=1\linewidth]{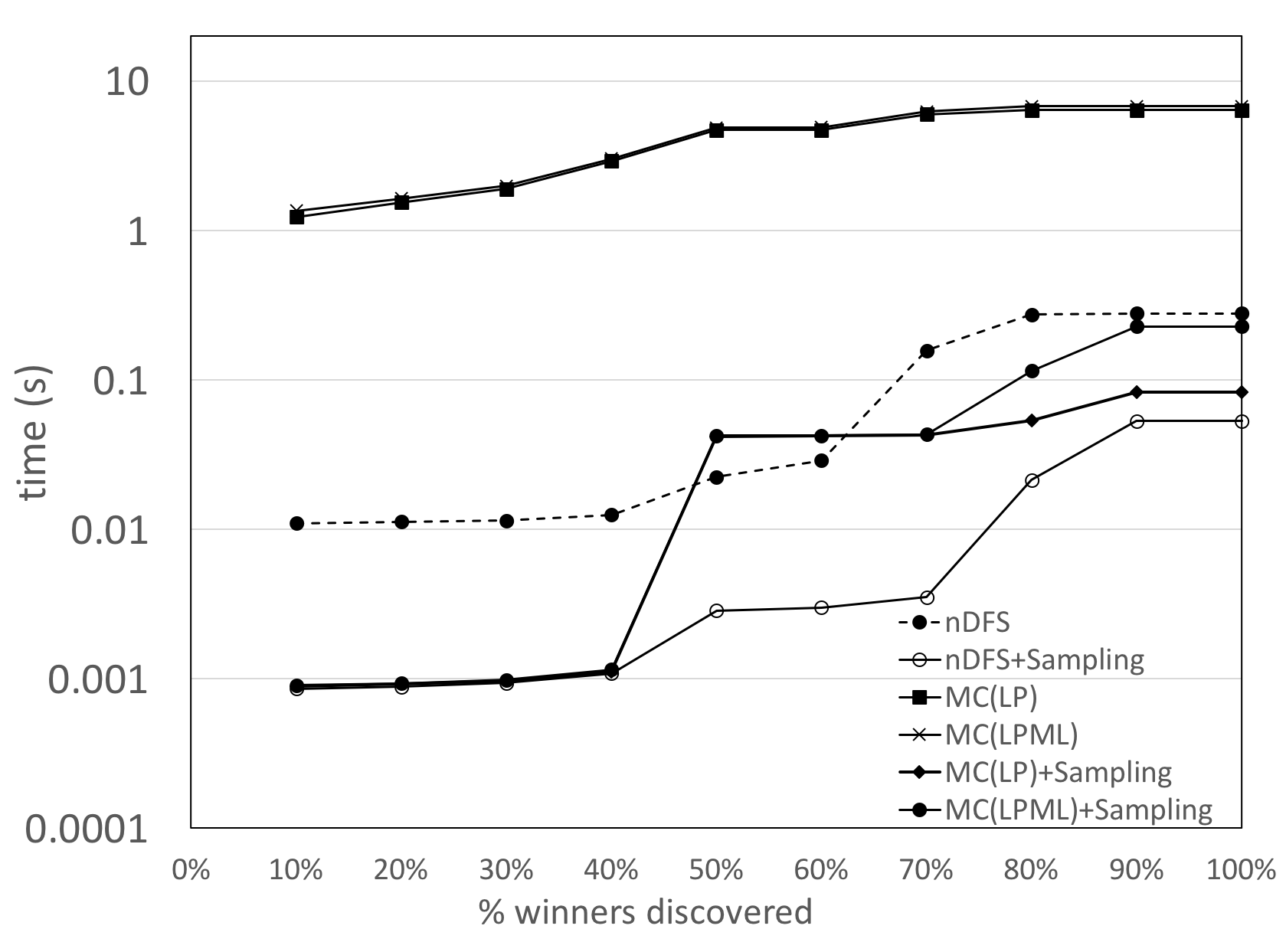}
	\caption{\label{fig:RP2} PUT-RP early discovery.} 
\end{figure}

We evaluate three algorithms together with their sampling variants for RP: \begin{enumerate*}[label=(\roman*)]\item {\bf  nDFS(LP):} naive DFS (Algorithm~\ref{algo:dfs}) with pruning  and local priority based on \# of candidate PUT-winners, \item {\bf MC(LP):}  maximal children based algorithm with SCC decomposition (Algorithm~\ref{algo:rpdfs}) and the same local priority as the previous one,  and \item {\bf MC(LPML):} its variant with a different local priority based on machine learning predictions.\end{enumerate*} 
Experimental results are summarized in Table~\ref{table:RP} and Figure~\ref{fig:RP2}. In short, the conclusion is that MC(LP) with sampling is the fastest algorithm for PUT-RP, and both the sampling variants of nDFS(LP) and MC(LP) perform well in early discovery. All comparisons are statistically significant with p-value$\leq$0.01 unless otherwise mentioned.


\noindent{\bf Pruning is Vital.}\label{sec:rpcp} Pruning plays a prominent part in the reduction of running time. 
Our contrasting experiment further justifies this argument: DFS without any pruning can only finish running on 531 profiles of our dataset before it gets stuck, taking 125.31 seconds in both running time and 100\%-discovery time on average, while DFS with pruning takes only 2.23 seconds and 2.18 seconds respectively with a surprising 50 times speedup.

\noindent{\bf Local Priority Improves Early Discovery.}\label{sec:rppriority}  In order to evaluate its efficacy, we test the naive DFS without local priority, and the average time for 100\%-discovery is 0.9976 seconds, which is about 3 times of that for nDFS(LP)'s 0.2778 seconds. Note that naive DFS without local priority is not shown in Figure~\ref{fig:RP2}. Local priority with machine learning (LPML) does not help as much as LP. For LPML,  we learn a neural network model using tenfold cross validation on 10,000 hard profiles and test on 1,000 hard profiles. The input features are the positional matrix, in- and out-degrees and plurality and Borda scores of all alternatives in the WMG.

\noindent{\bf Sampling is Efficient.}\label{sec:sampling} We test different algorithms with 200 samples and observe a large reduction in running time. The fastest combination is MC(LP) with sampling (in Table~\ref{table:RP}) with only 2.87s in running time and 0.08s in 100\%-discovery time on average. Sampling has the least improvement with p-value 0.02 when applied to nDFS.

\noindent{\bf Algorithms Perform Well on Real-World Data.} Using Preflib data, we find that MC(LP) performs better than naive DFS without local priority. We compare the two algorithms on 161 profiles with partial order. For MC(LP), the average running time and 100\%-discovery time are $1.33$s and $1.26$s, which have $46.0\%$ and $47.3\%$ reduction respectively compared to naive DFS. On 307 complete order profiles, the average running time and 100\%-discovery time of MC(LP) are both around 0.0179s with only a small reduction of 0.7\%, which is due to most profiles being easy cases without ties. In both experiments, we omit profiles with thousands of alternatives but very few votes which cause our machines to run out of memory.

\subsection{The Impact of The Size of Datasets on The Algorithms}\label{sec:sizeofdata} 
The sizes of $m$ and $n$ have different effects on searching space. Our algorithms can  deal with larger numbers of voters ($n$) without any problem. In fact, increasing $n$ reduces the likelihood of ties, which makes the computation easier. 

But for larger $m$,  the issue of memory constraint which comes from using cache to store visited states, becomes crucial. Without using cache, DFS becomes orders of magnitude slower. Our algorithm for PUT-STV with $m>30$ terminates with memory errors due to the exponential growth in state space, and our algorithm for PUT-RP is in a similar situation. Even with as few as $m = 10$ alternatives, the search space grows large. There are $ 3 ^ {\binom{m}{2}}$ possible states of the graph. For $m = 10$, this is $2.95\times10^{21}$ states. As such, due to memory constraints, currently we are only able to run our algorithms on profiles of size $m = n = 10$ for PUT-RP.

\section{Integer Linear Programming}\label{sec:ilp}

\noindent{\bf ILP for PUT-STV and Results.} The solutions correspond to the elimination of a single alternative in each of $m-1$ rounds and we test whether a given alternative is the PUT-winner by checking if there is a feasible solution when we enforce the constraint that the given alternative is not eliminated in any of the rounds. We omit the details due to the space constraint. Table~\ref{tbl:ilp_stv} summarizes the experimental results obtained using Gurobi's ILP solver. Clearly, the ILP solver takes far more time than even our most basic search algorithms without improvements.

\begin{table}[htp]
	\centering	
	\begin{tabular}[width=\linewidth]{|r|c|c|c|}\hline
		$m$& 10 & 20 & 30 \\
		$n$ & 10 & 20 & 30\\ \hline
		\# Profiles & 1000 & 2363 & 21\\
		Avg.~Runtime(s) & 1.14155 & 155.1874 & 12877.2792\\ \hline
	\end{tabular}
	\caption{\label{tbl:ilp_stv}ILP results for PUT-STV.}
\end{table}

\noindent{\bf ILP for PUT-RP.} We develop a novel ILP based on the characterization by Zavist and Tideman (Theorem~\ref{thm:iw}).
Let the {\em induced weight} (IW) between two vertices $a$ and $b$ be the maximum path weight over all paths from $a$ to $b$ in the graph. The path weight is defined as the minimum edge weight of a given path. An edge $(u,v)$ is {\em consistent} with a ranking $R$ if $u$ is preferred to $v$ by $R$. $G_R$ is a graph whose vertices are $\mathcal A$ and whose edges are exactly every edge in $\text{wmg}_{\geq0}(P)$ consistent with a ranking $R$. Thus there is a topological ordering of $G_R$ that is exactly $R$.

\begin{ex} In Figure~\ref{fig:rpex}, consider the induced weight from $D$ to $A$ in the bottom left graph. There are three distinct paths: $P_1 = \{D \rightarrow A\}$, $P_2 = \{D \rightarrow C \rightarrow A\}$, and $P_3 = \{D \rightarrow C \rightarrow B \rightarrow A\}$. The weight of $P_1$, or $W(P_1) = 1$, $W(P_2) = 3$ and $W(P_3) = 1$. Thus, IW$(D,A) = 3$, and note that IW$(D,A) \geq w_{(A,D)} = -1$.
\end{ex}

\begin{thm}~\cite{Zavist1989} \label{thm:iw}
	For any profile $P$ and for any strict ranking $R$, the ranking $R$ is the outcome of the ranked pairs procedure if and only if $G_R$ satisfies the following property for all candidates $i,j \in \mathcal A$: $\forall i \succ_R j, \text{ IW}(i,j) \geq w_{(j,i)}$.
\end{thm}

Based on Theorem~\ref{thm:iw}, we provide a novel ILP formulation of the PUT-RP problem. See Appendices~\ref{sec:ilpapp} for details.

\noindent{\bf Results.} Out of $1000$ hard profiles, the RP ILP ran faster than DFS on $16$ profiles. On these $16$ profiles, the ILP took only $41.097\%$ of the time of the DFS to compute all PUT-winners on average. However over all $1000$ hard profiles, DFS is much faster on average: 29.131 times faster. We propose that on profiles where DFS fails to compute all PUT-winners, or for elections with a large number of candidates, we can fall back on the ILP to solve PUT-RP.
\section{Future Work}\label{sec:cls}

There are many other strategies we wish to explore. For the heuristic function, there are more local priority functions we could test based on exploiting specific structures of the voting rules to encourage early discovery. Further machine learning techniques or potentially reinforcement learning could prove useful here. For PUT-RP, we want to specifically test the performance of our SCC-based algorithm on large profiles with many SCCs, since currently our dataset contains a low proportion of multi-SCC profiles. Also, we want to extend our search algorithm  to multi-winner voting rules like the Chamberlin--Courant rule, which is known to be NP-hard to compute an optimal committee for general preferences \cite{Procaccia:Multi}.

\section{Acknowledgments}
We thank all anonymous reviewers for helpful comments and suggestions. This work is supported by NSF \#1453542 and ONR \#N00014-17-1-2621.

\appendix
\section{Appendices}\label{stylefiles}
\subsection{Proof of Theorem 1}
	Suppose $H=(\mathcal A, E)$ is composed of SCCs $S_i\;(i=1,\cdots,k)$, and the set of bridge edges $B$.
	We need to prove $C=(\mathcal A, B\cup\bigcup_{i=1}^{k}C_i)$ is a maximal child of $H$, where $C_i$ is one of the maximal children of $S_i$. Suppose for contradiction that  $C$ is not one maximal child of $H$. From Lemma~\ref{lem:scc} we have that $C$ is acyclic.
	then it suffice that $\exists\; e\in E\setminus \left(B\cup\bigcup_{i=1}^{k}C_i \right)=\left(B\cup\bigcup_{i=1}^{k}S_i\right)\setminus \left(B\cup\bigcup_{i=1}^{k}C_i\right)
	=\bigcup_{i=1}^{k}S_i\setminus C_i$, s.t.  $\{e\}\cup B\cup\bigcup_{i=1}^{k}C_i$ is still acyclic.
	
	W.l.o.g., let's assume $e\in S_1\setminus C_1$. So $\{e\}\cup C_1$ is acyclic. But this contradicts the fact that $C_1$ is a maximal child of $S_1$, since by definition, $\forall$ $e'\in S_1\setminus C_1 $, $C_1\cup\{e'\}$ is cyclic. \qed

\begin{lem}\label{lem:scc}
	For any directed graph $H=(\mathcal A,E)$, composed of strongly connected components $S_i\;(i=1,\cdots,k)$, and the set of bridge edges $B$, the directed graph $G=(\mathcal A, B\cup\bigcup_{i=1}^{k}C_i)$ is also acyclic, where $C_i$ is one of the maximal children of $S_i$. 
\end{lem}

\subsection{ILP Formulation for PUT-RP}\label{sec:ilpapp}
We can test whether a given alternative $i^*$ is a PUT-RP winner if there is a solution subject to the constraint that there is no path from any other alternative to $i^*$. The variables are: \begin{enumerate*}[label=(\roman*)]
	\item A binary indicator variable $X_{i,j}^t$ of whether there is an $i\to j$ path using locked in edges from $\bigcup T_{i\le t}$, for each $i,j\le m, t\le K$.
	\item A binary indicator variable $Y_{i,j,k}^t$ of whether there is an $i\to k$ path involving node $j$ using locked in edges from tiers $\bigcup T_{i\le t}$, for each $i,j,k\le m, t\le K$.
\end{enumerate*}

We can determine all PUT-winners by selecting every alternative $i^* \leq m$, adding the constraint $\sum_{j\le m, j\neq i^*} X_{j,i^*}^K = 0$, and checking the feasibility with the constraints below:\\
\noindent$\bullet$~To enforce Theorem~\ref{thm:iw}, for every pair $i, j \le m$, such that $(j,i) \in T_t$, we add the constraint $X_{i,j}^{t} \ge X_{i,j}^K$.\\
\noindent$\bullet$~In addition, we have constraints to ensure that \begin{enumerate*}[label=(\roman*)]\item locked in edges from $\bigcup_{t\le K} T_t$ induce a total order over $\mathcal A$ by enforcing asymmetry and transitivity constraints on $X_{i,j}^K$ variables, and \item enforcing that if $X_{i,j}^t=1$, then $X_{i,j}^{\hat t >t}=1$.\end{enumerate*}\\
\noindent$\bullet$~Constraints ensuring maximum weight paths are selected:
\begin{equation}\nonumber
	\begin{aligned}
	\left.~~~\begin{tabular}{l}
	$\forall i,j,k\le m, t\le K$,\\
	$Y_{i,j,k}^t \ge X_{i,j}^t + X_{j,k}^t - 1$\\
	$Y_{i,j,k}^t \le\frac{X_{i,j}^t + X_{j,k}^t}{2}$
	\end{tabular}\right\}& \text{$i\to j \to k$}\\
	\end{aligned}
\end{equation}
\begin{equation}\nonumber
	\begin{aligned}
	\left.\begin{tabular}{l}
	$\forall i,k\le m, t \le K,$\\ 
	if $(i,k) \in E^{\hat t \le t}$, $X_{i,k}^t \ge X_{i,k}^K$
	\end{tabular}\right\}& \text{$(i,k)$}\\
	\end{aligned}
\end{equation}	
	\begin{equation}\nonumber
	\begin{aligned}
	\left.\begin{tabular}{p{2.4cm}l}
	& $\forall j \le m$, \\
	& $X_{i,k}^t \ge Y_{i,j,k}^t$,\\
	if $(i,k) \in T_{\hat t > t}$, & $X_{i,k}^t\le \sum_{j\le m}Y_{i,j,k}^t$,\\
	if $(i,k) \in T_{\hat t \le t}$, & $X_{i,k}^t\le \sum_{j\le m}Y_{i,j,k}^t + X_{i,k}^K$\\
	\end{tabular}\right\} & \text{$i\to k$}\\
	\end{aligned}
\end{equation}


\bibliographystyle{aaai}

\end{document}